\documentclass{article}
\usepackage{spconf,amsmath,graphicx}
\usepackage{subfigure}
\usepackage{multirow}
\usepackage{url}
\graphicspath{{./figure/}}
\newcommand{\vsSectionUp}{\vspace{-3.5mm}}
\newcommand{\vsSectionDown}{\vspace{-2mm}}

\newcommand{\vsTableUp}{\vspace{-3mm}}
\newcommand{\vsTableMiddle}{\vspace{-3mm}}
\newcommand{\vsTableDown}{\vspace{-2mm}}
\newcommand{\vsFigureUp}{\vspace{-2mm}}

\newcommand{\vsFigureMiddle}{\vspace{-6mm}}
\newcommand{\vsFigureDown}{\vspace{-2mm}}
\newcommand{\vsEquationUp}{\vspace{-2mm}}
\newcommand{\vsEquationDown}{\vspace{-2mm}}
\title{On Classification of Distorted Images with Deep Convolutional Neural Networks}
\name{Yiren Zhou, Sibo Song, Ngai-Man Cheung}
\address{Singapore University of Technology and Design}
\begin{document}
\ninept
\maketitle
\vsSectionUp
\begin{abstract}
\vsSectionDown

Image blur and image noise are common distortions during image acquisition.
In this paper, 
we systematically study the effect of image distortions on the deep neural network (DNN) image classifiers.  First, we examine the DNN classifier performance under four types of distortions.  Second, we propose two approaches to alleviate the effect of image distortion: re-training and fine-tuning with noisy images. Our results suggest that, under certain conditions, fine-tuning with noisy images can alleviate much effect due to distorted inputs, and is more practical than re-training.



\end{abstract}
\vsSectionUp
\section{Introduction}
\vsSectionDown

Recently, deep neural networks~(DNNs) have achieved superior results on many computer vision tasks~\cite{razavian2014cnn}.  In image classification, 
DNN approaches such as Alexnet~\cite{krizhevsky2012imagenet} have significantly improved the accuracy compared to previously hand-crafted features. Further works on DNN~\cite{Simonyan14c,he2015deep} continue to advance the DNN structures and improve the performance.


In practical applications, various types of distortions may occur in the captured images.  For example, images captured with moving cameras may suffer from motion blur.  In this paper, we systematically study the effect of image distortion on DNN-based image classifiers.  We also examine some strategy to alleviate the impact of image distortion on the classification accuracy.

Two main categories of image distortions are image blur and image noise~\cite{buades2005review}. They are caused by various issues during image acquisition.  For example, defocus blur occurs when the camera is out of focus.  Motion blur is caused by relative movement between the camera and the view, which is common for smartphone-based image analysis~\cite{hossein:spm:2016,victor:icip:2016,toan:embc:2014}.  Image noise is usually caused by poor illumination and/or high temperature, which degrade the performance of the charge coupled device~(CCD) inside the camera.

When we apply a DNN classifier in a practical application, it is  possible that some image blur and noise would occur in the input images.
These degradations would affect the performance of the DNN classifier. Our work makes several contributions to this problem.  First, we study the effect of image distortion on the DNN classifier.  We examine the DNN classifier performance under four types of distortions on the input images: motion blur, defocus blur, Gaussian noise and a combination of them.    Second, we examine two approaches to alleviate the effect of image distortion.  In one approach, we re-train the whole network with the noisy images. We find that this approach can improve the accuracy when classifying distorted images.  However, re-training requires large training datasets for very deep networks.  
Inspired by~\cite{yosinski2014transferable},
in another approach, we fine-tune the first few layers of the network with distorted images.
Essentially, we adjust the low-level filters of the DNN to match the characteristics of the distorted images. 

Some previous works have studied the effect of image distortion~\cite{yiren:tcsvt:2016}.  Focusing on DNN, 
Basu et al.~\cite{basu2015learning} proposed a new 
model modified from deep belief nets to deal with noisy inputs.
They reported good results on a noisy dataset called n-MNIST, which contains Gaussian noise, motion blur, and reduced contrast compared to original MNIST dataset. Recently,
Dodge and Karam~\cite{dodge2016understanding}  reported the degradation due to various image distortions in several DNN.  
Compared to these works, we perform  a unified study to investigate effect of image distortion on (i) hand-written digit classification and (ii) natural image classification.  Moreover, we examine using re-training and fine-tuning with noisy images to alleviate the effect.

In classification of ``clean'' images (i.e., without distortion), 
some previous work has attempted to introduce noise to the training data~\cite{rifai2011adding, yixinluo2014noise}.  In these works, their purpose is to use noise to regularize the model in order to prevent overfitting during training.
On the contrary, our goal is to understand the benefits of using noisy training data in classification of distorted images.  Our results also suggest that, under certain conditions, fine-tuning using noisy images can be an effective and practical approach.
\vsSectionUp
\section{Deep architecture}
\vsSectionDown

In this section, we briefly introduce the idea of deep neural network~(DNN). There are many types of DNN, here we mainly introduce deep convolutional neural network~(DCNN), a detailed introduction for DNN can be found in~\cite{dl2016docu}.

DNN is a machine learning architecture that is inspired by humans' central nervous systems. The basic element in DNN is neuron. In DNN, neighborhood layers are fully connected by neurons, and one DNN can have multiple concatenated layers. Those layers together form a DNN.




DNN has achieved great performance for problems on small images~\cite{hinton2006reducing}. However, for problems with large images, conventional DNN need to use all the nodes in the previous layer as inputs to the next layer, and this lead to a model with a very large number of parameters, and impossible to train with a limited dataset and computation sources. The idea of convolutional neural network~(CNN) is to make use of the local connectivity of images as prior knowledge, that a node is only connected to its neighborhood nodes in the previous layer. This constraint significantly reduces the size of the model, while preserving the necessary information from an image. 

For a convolutional layer, each node is connected to a local region in the input layer, which is called receptive field. All these nodes form an output layer. For all these nodes in the output layer, they have different kernels, but they share the same weights when calculating activation function.

\begin{figure}
\vsFigureUp
\centering
\includegraphics[width=\columnwidth]{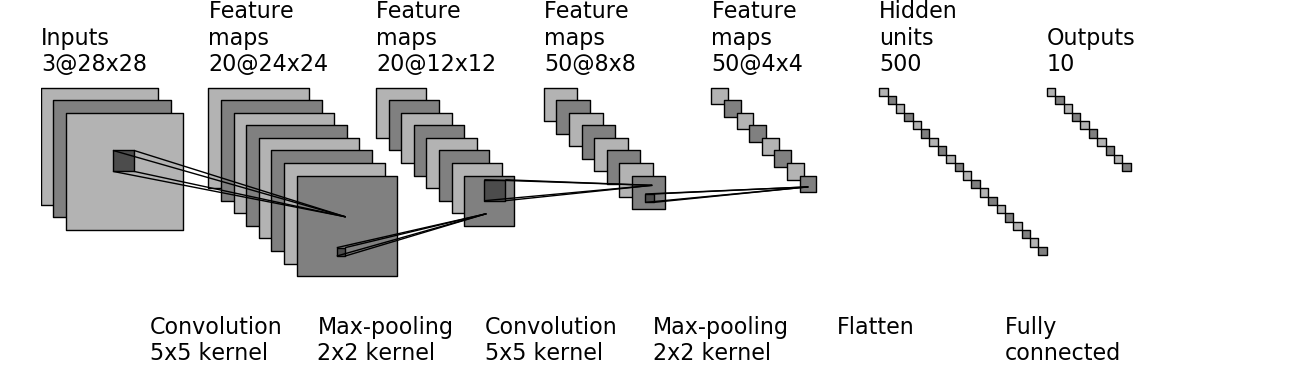}
\vsFigureMiddle
\caption[]{Structure of LeNet-5.}
\label{fig:lenet}
\end{figure}

\begin{figure}
\vsFigureUp
\centering
\includegraphics[width=\columnwidth]{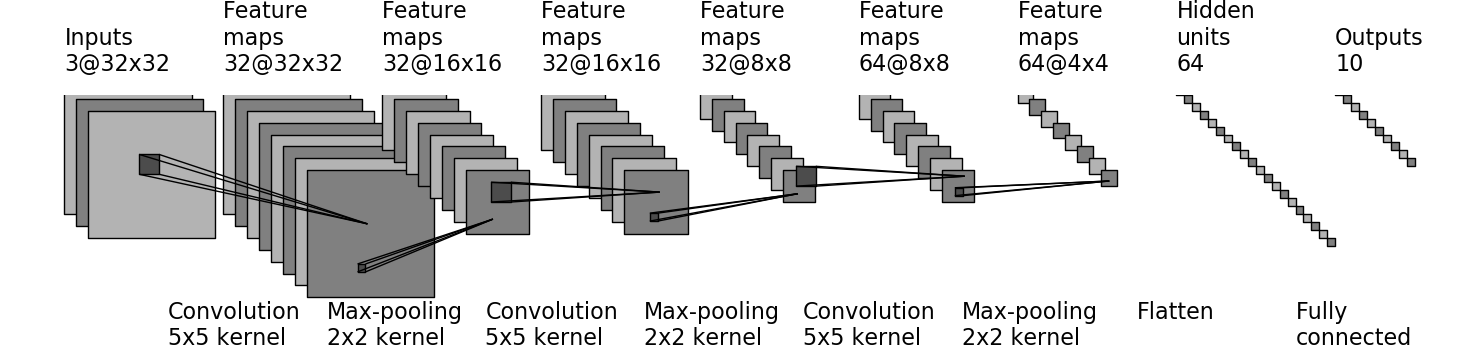}
\vsFigureMiddle
\caption[]{Structure of CIFAR10-quick model.}
\label{fig:cifar10-quick}
\vsFigureDown
\end{figure}

Fig.~\ref{fig:lenet} shows the architecture of LeNet-5, which is used for digit image classification on MNIST dataset~\cite{lecun1998gradient}.
From the figure we can see that the model has two convolutional layers and their corresponding pooling layers. This is the convolutional part for the model. The following two layers are flatten and fully connected layers, these layers are inherited from conventional DNN.


%
\vsSectionUp
\section{Experimental settings}
\vsSectionDown

We conduct experiment on both relatively small datasets~\cite{lecun1998gradient,krizhevsky2009learning} and a large image dataset, ImageNet~\cite{imagenet_cvpr09}.  We examine different full training / fine-tuning configurations on some small datasets to gain insight into their effectiveness.  We then examine and validate our approach on ImageNet dataset.


We conduct the experiment using MatConvNet~\cite{vedaldi15matconvnet}, a MATLAB toolbox which can run and learn convolutional neural networks. All the experiments are conducted on a Dell T5610 WorkStation with Intel Xeon E5-2630 CPU.

\begin{figure}[htbp]
\vsFigureUp
\subfigure{\rotatebox{90}{Motion}\\\rotatebox{90}{blur}}
\subfigure{\includegraphics[width=0.16\columnwidth]{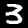}}
\subfigure{\includegraphics[width=0.16\columnwidth]{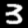}}
\subfigure{\includegraphics[width=0.16\columnwidth]{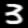}}
\subfigure{\includegraphics[width=0.16\columnwidth]{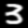}}
\subfigure{\includegraphics[width=0.16\columnwidth]{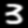}}
\\
\subfigure{\rotatebox{90}{Defocus}\\\rotatebox{90}{blur}}
\subfigure{\includegraphics[width=0.16\columnwidth]{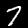}}
\subfigure{\includegraphics[width=0.16\columnwidth]{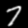}}
\subfigure{\includegraphics[width=0.16\columnwidth]{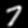}}
\subfigure{\includegraphics[width=0.16\columnwidth]{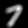}}
\subfigure{\includegraphics[width=0.16\columnwidth]{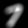}}
\\
\subfigure{\rotatebox{90}{Gaussian}\\\rotatebox{90}{noise}}
\subfigure{\includegraphics[width=0.16\columnwidth]{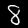}}
\subfigure{\includegraphics[width=0.16\columnwidth]{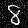}}
\subfigure{\includegraphics[width=0.16\columnwidth]{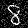}}
\subfigure{\includegraphics[width=0.16\columnwidth]{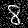}}
\subfigure{\includegraphics[width=0.16\columnwidth]{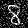}}
\\
\subfigure{\rotatebox{90}{All}\\\rotatebox{90}{combine}}
\subfigure{\includegraphics[width=0.16\columnwidth]{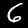}}
\subfigure{\includegraphics[width=0.16\columnwidth]{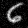}}
\subfigure{\includegraphics[width=0.16\columnwidth]{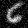}}
\subfigure{\includegraphics[width=0.16\columnwidth]{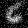}}
\subfigure{\includegraphics[width=0.16\columnwidth]{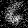}}
\\
\vsFigureMiddle
\caption[]{Example MNIST images after different amount of motion blur, defocus blur, Gaussian noise, and all combined.}
\label{fig:example-mnist}
\vsFigureDown
\end{figure}


\textbf{Deep architectures and datasets:}
In this evaluation we consider three well-known dataset: MNIST~\cite{lecun1998gradient}, CIFAR-10~\cite{krizhevsky2009learning}, and ImageNet~\cite{imagenet_cvpr09}.

MNIST is a handwritten digits dataset with 60000 training images and 10000 test images. Each image is a $28\times28$ greyscale image, belonging to one digit class from '0' to '9'. For MNIST, we use LeNet-5~\cite{lecun1998gradient} for classification. The structure of LeNet-5 we use is shown in Fig.~\ref{fig:lenet}. This network has 6 layers and 4 of them have parameters to train: the first two convolutional layers, flatten and fully connected layers.

We consider two approaches to deal with distorted images: fine-tuning and re-training with noisy images.

In fine-tuning, we start from the pre-trained model trained with the original dataset (i.e., images without distortion).  We fine-tune the first N layers of the model on a distorted dataset while fixing the parameters in the remaining layers. The reason to fix parameters in the last layers is that image blur and noise are considered to have more effect on low-level features in images, such as color, edge, and texture features. However, these distortions have little effect on high-level information, such as the semantic meanings of an image~\cite{liu2015deep}. Therefore, in fine-tuning, we focus on the starting layers of a DNN, which contain more low-level information.
As an example, 
for LeNet-5 we have 4 layers with parameters, that means N is ranging from 1 to 4. We denote fine-tuning methods as first-1 to first-4. 

In re-training, we train the whole network with the distorted dataset from scratch and do not use the pre-trained model. We denote the re-training method as re-training.

For re-training LeNet-5, we set the learning rate to $10^{-3}$, and the number of epochs to 20. For fine-tuning, we set learning rate to $10^{-5}$~($1\%$ of the re-training learning rate), and number of epochs to 15. Each epoch takes about 1 minute, so the training procedure takes about 20 minutes for re-training, and 15 minutes for fine-tuning.

CIFAR-10 dataset consists of 60000 $32\times32$ color images in 10 classes, with 6000 images per class. 50000 are training images, and 10000 are test images. To make the training faster, we use a fast model provided in MatConvNet~\cite{vedaldi15matconvnet}. The structure of CIFAR10-quick model is shown in Fig.~\ref{fig:cifar10-quick}.

Similar to previous approaches for MNIST, we use fine-tuning and re-training for CIFAR distorted dataset. There are 5 layers with parameters in CIFAR10-quick model, so we have first-1 to first-5 as fine-tuning methods. The re-training method is denoted as re-training.

For re-training CIFAR10-quick, we set the number of epochs to 45. Learning rate is set to $5\times10^{-2}$ for first 30 epochs, $5\times10^{-3}$ for the following 10 epochs, and $5\times10^{-4}$ for the last 5 epochs. For fine-tuning, we set the number of epochs to 30. Learning rate is $5\times10^{-4}$ for first 25 epochs, and $5\times10^{-5}$ for last 5 epochs. Each epoch takes about 3 minutes, so the training procedure takes about 135 minutes for re-training, and 90 minutes for fine-tuning.

\begin{figure}[htbp]
\vsFigureUp
\subfigure{\rotatebox{90}{Motion}\\\rotatebox{90}{blur}}
\subfigure{\includegraphics[width=0.16\columnwidth]{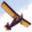}}
\subfigure{\includegraphics[width=0.16\columnwidth]{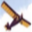}}
\subfigure{\includegraphics[width=0.16\columnwidth]{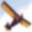}}
\subfigure{\includegraphics[width=0.16\columnwidth]{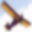}}
\subfigure{\includegraphics[width=0.16\columnwidth]{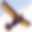}}
\\
\subfigure{\rotatebox{90}{Defocus}\\\rotatebox{90}{blur}}
\subfigure{\includegraphics[width=0.16\columnwidth]{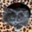}}
\subfigure{\includegraphics[width=0.16\columnwidth]{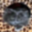}}
\subfigure{\includegraphics[width=0.16\columnwidth]{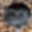}}
\subfigure{\includegraphics[width=0.16\columnwidth]{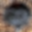}}
\subfigure{\includegraphics[width=0.16\columnwidth]{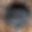}}
\\
\subfigure{\rotatebox{90}{Gaussian}\\\rotatebox{90}{noise}}
\subfigure{\includegraphics[width=0.16\columnwidth]{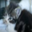}}
\subfigure{\includegraphics[width=0.16\columnwidth]{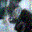}}
\subfigure{\includegraphics[width=0.16\columnwidth]{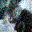}}
\subfigure{\includegraphics[width=0.16\columnwidth]{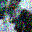}}
\subfigure{\includegraphics[width=0.16\columnwidth]{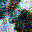}}
\\
\subfigure{\rotatebox{90}{All}\\\rotatebox{90}{combine}}
\subfigure{\includegraphics[width=0.16\columnwidth]{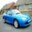}}
\subfigure{\includegraphics[width=0.16\columnwidth]{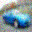}}
\subfigure{\includegraphics[width=0.16\columnwidth]{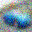}}
\subfigure{\includegraphics[width=0.16\columnwidth]{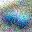}}
\subfigure{\includegraphics[width=0.16\columnwidth]{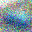}}
\\
\vsFigureMiddle
\caption[]{Example CIFAR-10 images after different amount of motion blur, defocus blur, Gaussian noise, and all combine.}
\label{fig:example-cifar}
\vsFigureDown
\end{figure}

Here we also present an evaluation on ILSVRC2012 dataset. ILSVRC2012~\cite{ILSVRC15} is a large-scale natural image dataset containing more than one million images in 1000 categories. The images and categories are selected from ImageNet~\cite{imagenet_cvpr09}. To understand the effect of limited data in many applications, we randomly choose 50000 images from training dataset for training, and use validation set of ILSVRC2012, which contains 50000 images, for testing.

We use fine-tuning method for ILSVRC2012 validation set with a pre-trained Alexnet model~\cite{krizhevsky2012imagenet}.
We do not use re-training method here, because re-training Alexnet using only small part of the training set of ILSVRC2012 would cause overfitting.
We fine-tune the first 3 layers of Alexnet, while fixing the remaining layers. For fine-tuning process, the number of epochs is set to 20. The learning rate is set to $10^{-8}$ to $10^{-10}$ from epoch 1 to epoch 20, decreases by log space. We also use a weight decay of $5\times10^{-4}$. Approximate training time is 90 minutes for each epoch, and 30 hours for total process.

Regarding  the computation time, fine-tuning takes less time than re-training on the MNIST and CIFAR-10 dataset. For ILSVRC2012 validation set, we also need to use fine-tuning method in order to prevent overfitting.

\begin{figure}[htbp]
\vsFigureUp
\centering
\subfigure[]{
    \includegraphics[width=0.45\columnwidth]{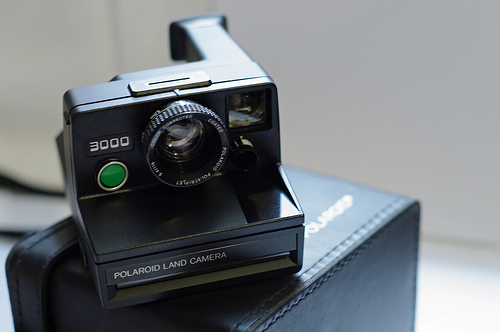}
    \label{fig:original-imagenet}
}
\subfigure[]{
    \includegraphics[width=0.45\columnwidth]{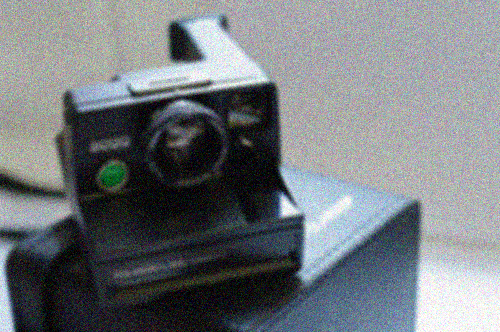}
    \label{fig:noise-imagenet}
}\\
\vsFigureMiddle
\caption[]{Example images from ImageNet validation set. (a) is the original image. (b) is the distorted image.}
\label{fig:example-imagenet}
\vsFigureDown
\end{figure}


\textbf{Types of blur and noise:}
In this experiment, we consider two types of blur: motion blur and defocus blur, and one type of noise: Gaussian noise.

Motion blur is a typical type of blur usually caused by camera shaking and/or fast-moving of the photographed objects. We generate the motion blur kernel using random walk~\cite{hradivs2015convolutional}. For each step size, we move the motion blur kernel in a random direction by 1-pixel. The size of the motion blur kernel is sampled from $[0,4]$.

Defocus blur happens when the camera loses focus of an image. We generate the defocus blur by uniform anti-aliased disc. The radius of the disc is sampled from $[0,4]$.

After generating a motion or a defocus blur kernel for one image, we use this kernel for convolution operation on the whole image to generate a blurred image. 

Gaussian noise is caused by poor illumination and/or high temperature, which prevents CCD in a camera from getting correct pixel values. We choose Gaussian noise with zero means, and with standard deviation $\sigma$ sampled from $[0,4]$ on a color image with an integer value in $[0,255]$.

Finally, we consider a combination of all the above three types of distortions. The value of each noise is sampled from $[0,4]$, respectively.

\begin{figure*}[!t]
\vsFigureUp
\subfigure[]{
      \includegraphics[width=0.45\columnwidth]{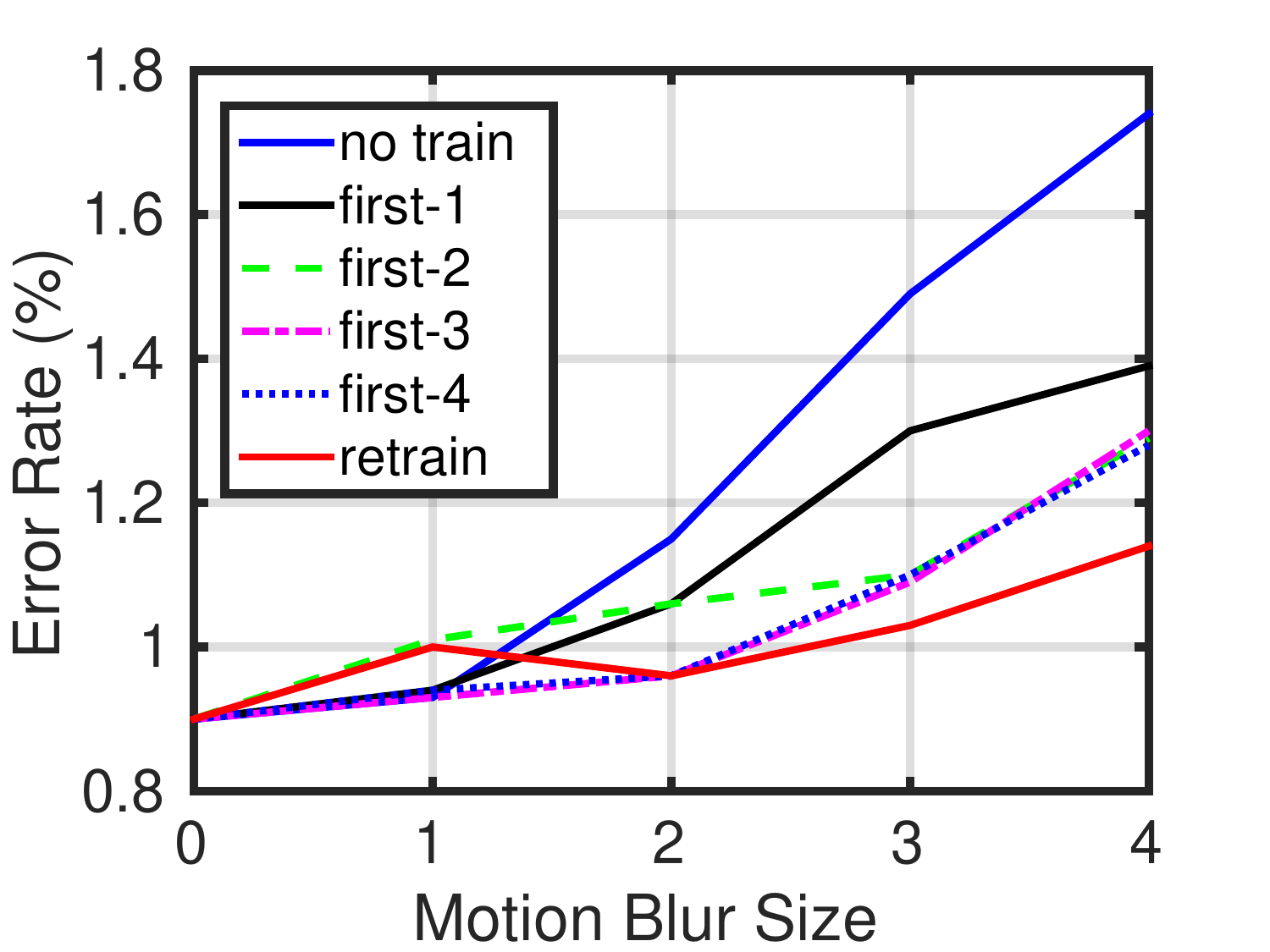}
       \label{fig:mnist-motion}
}
\subfigure[]{
      \includegraphics[width=0.45\columnwidth]{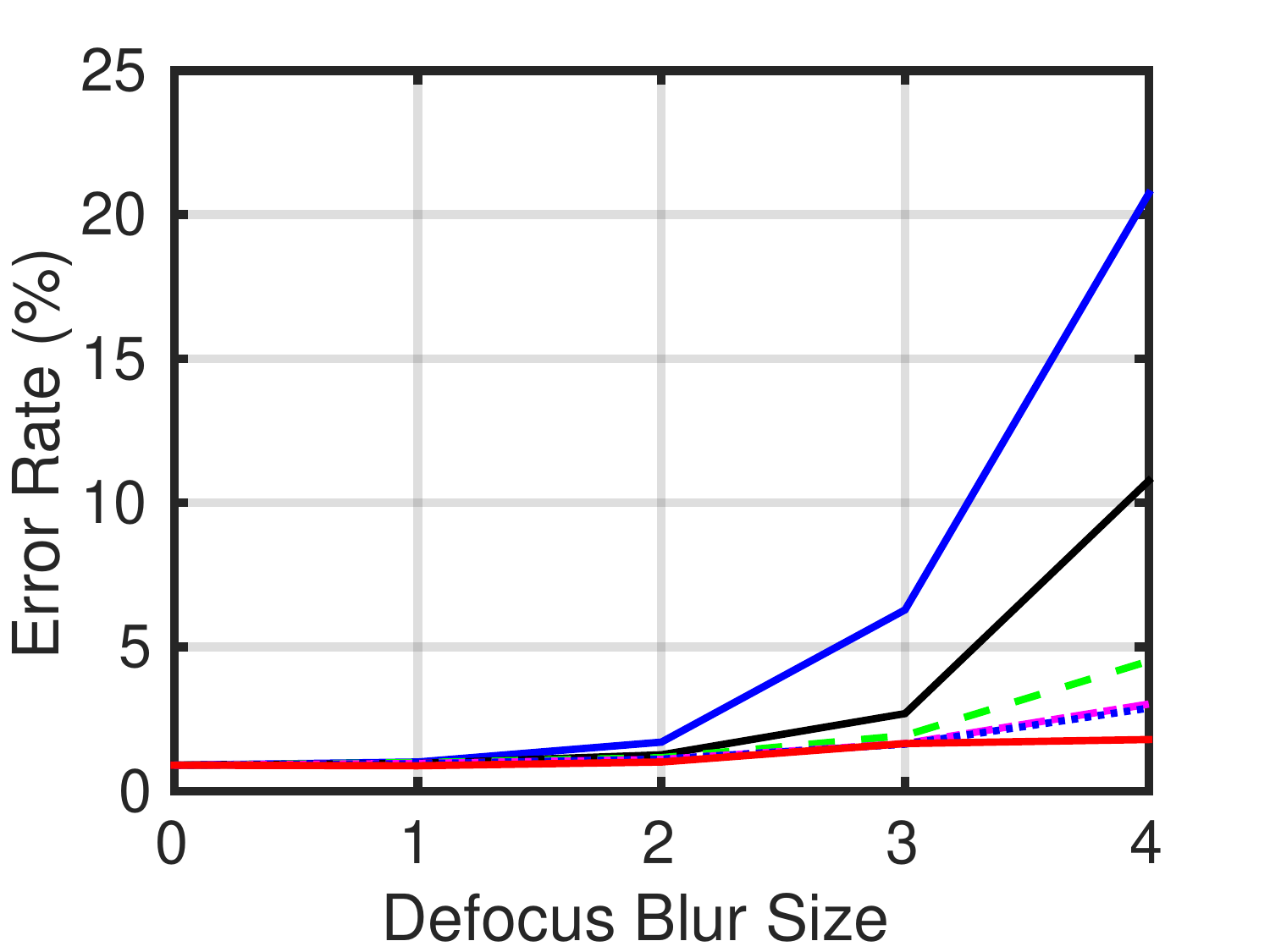}
       \label{fig:mnist-defocus}
}
\subfigure[]{
      \includegraphics[width=0.45\columnwidth]{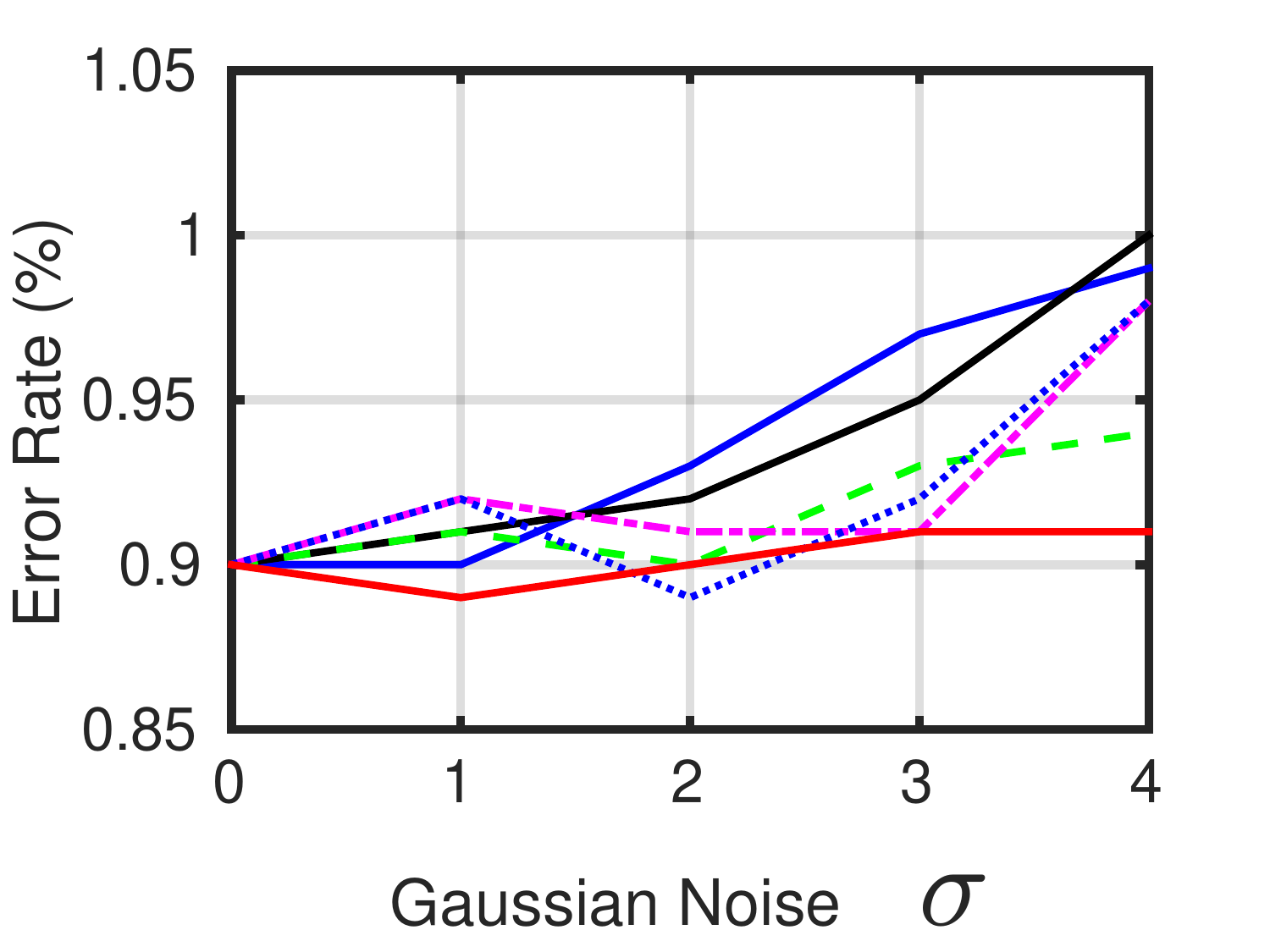}
       \label{fig:mnist-gaussian}
}
\subfigure[]{
      \includegraphics[width=0.45\columnwidth]{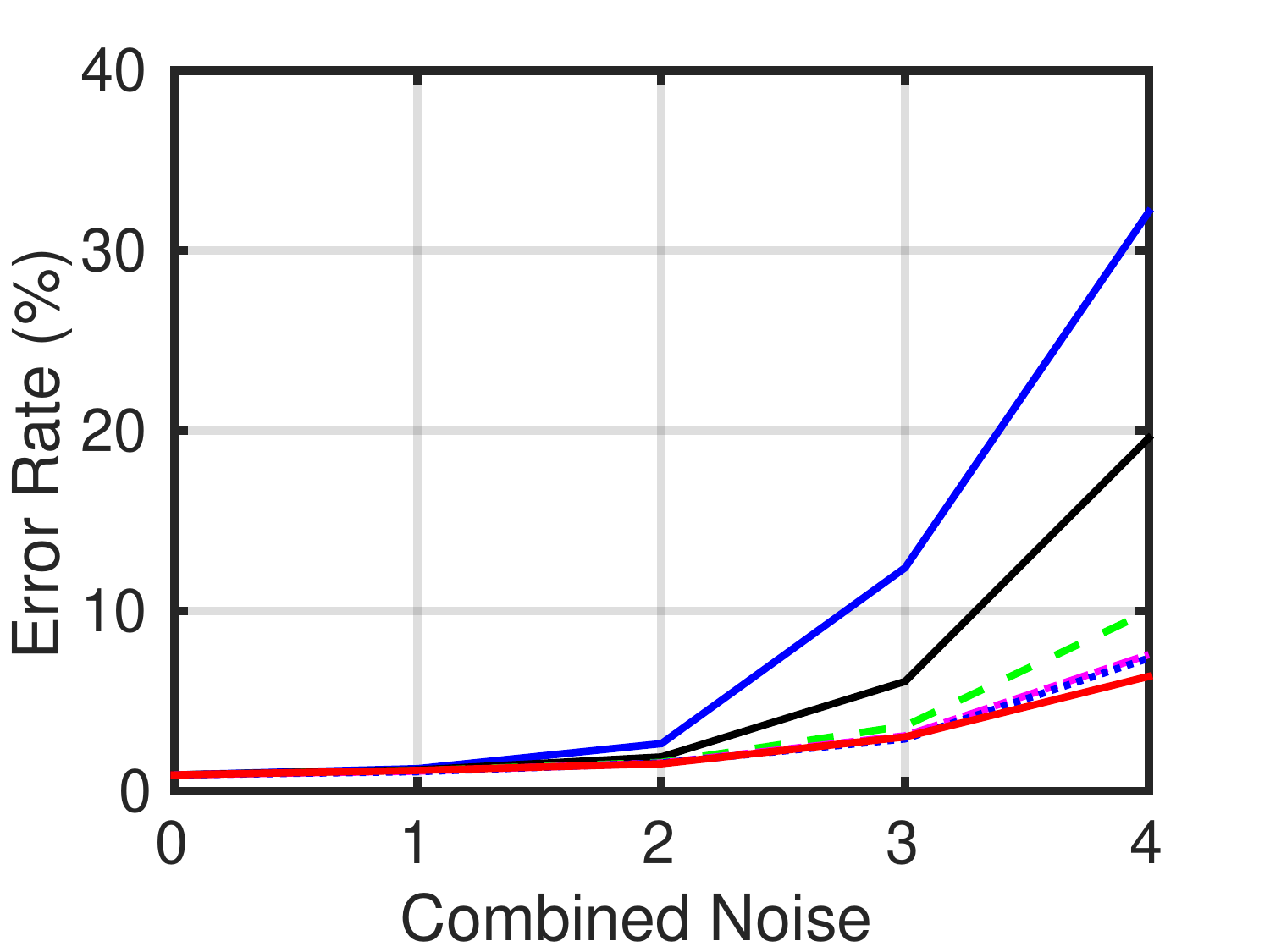}
       \label{fig:mnist-all}
}\\
\vsFigureMiddle
\caption[]{Error rates for LeNet-5 model on MNIST dataset under different blurs and noises.}
\label{fig:err-mnist}
\vsFigureDown
\end{figure*}

\begin{figure*}[!t]
\vsFigureUp
\subfigure[]{
      \includegraphics[width=0.45\columnwidth]{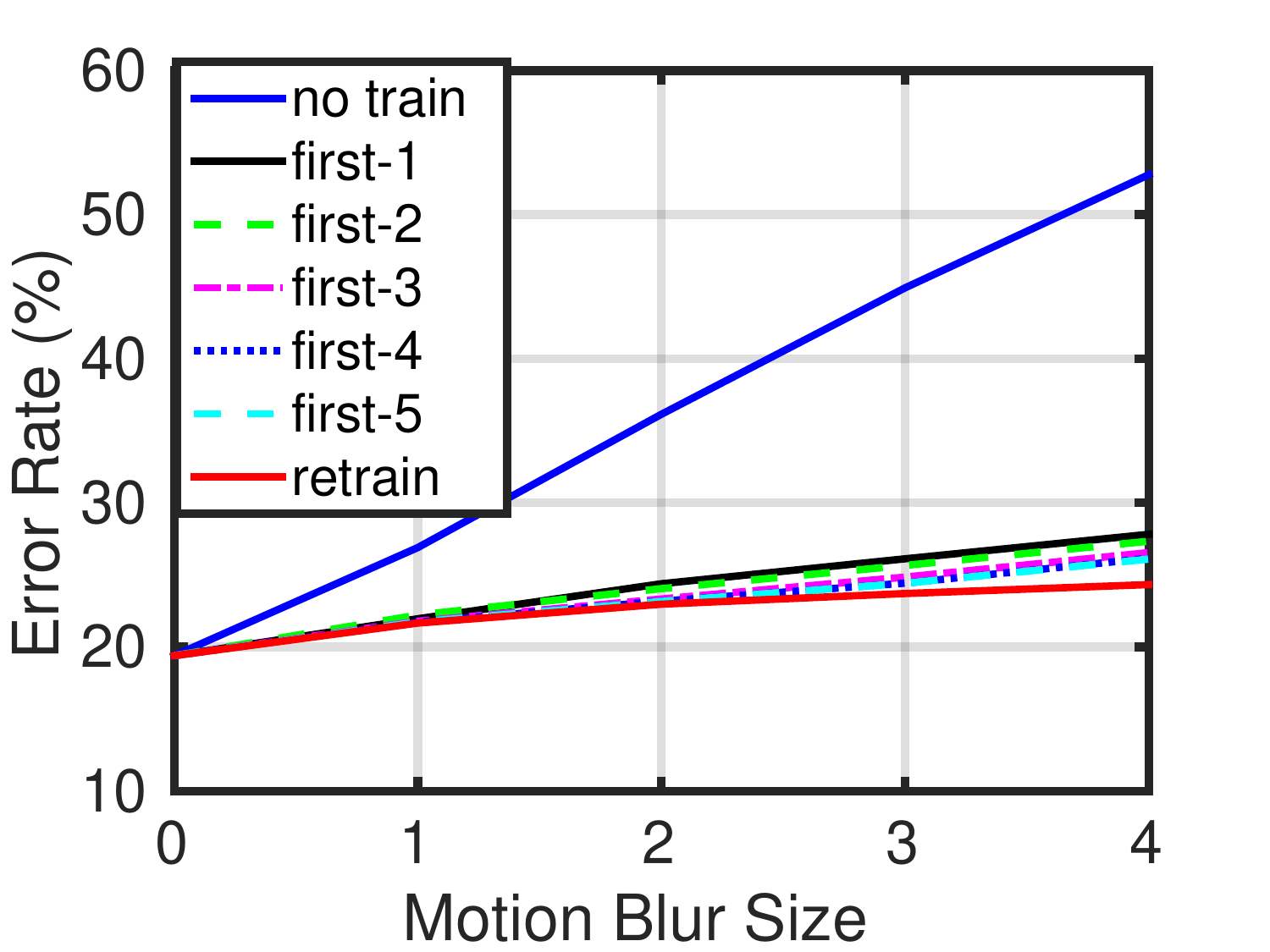}
       \label{fig:cifar-motion}
}
\subfigure[]{
      \includegraphics[width=0.45\columnwidth]{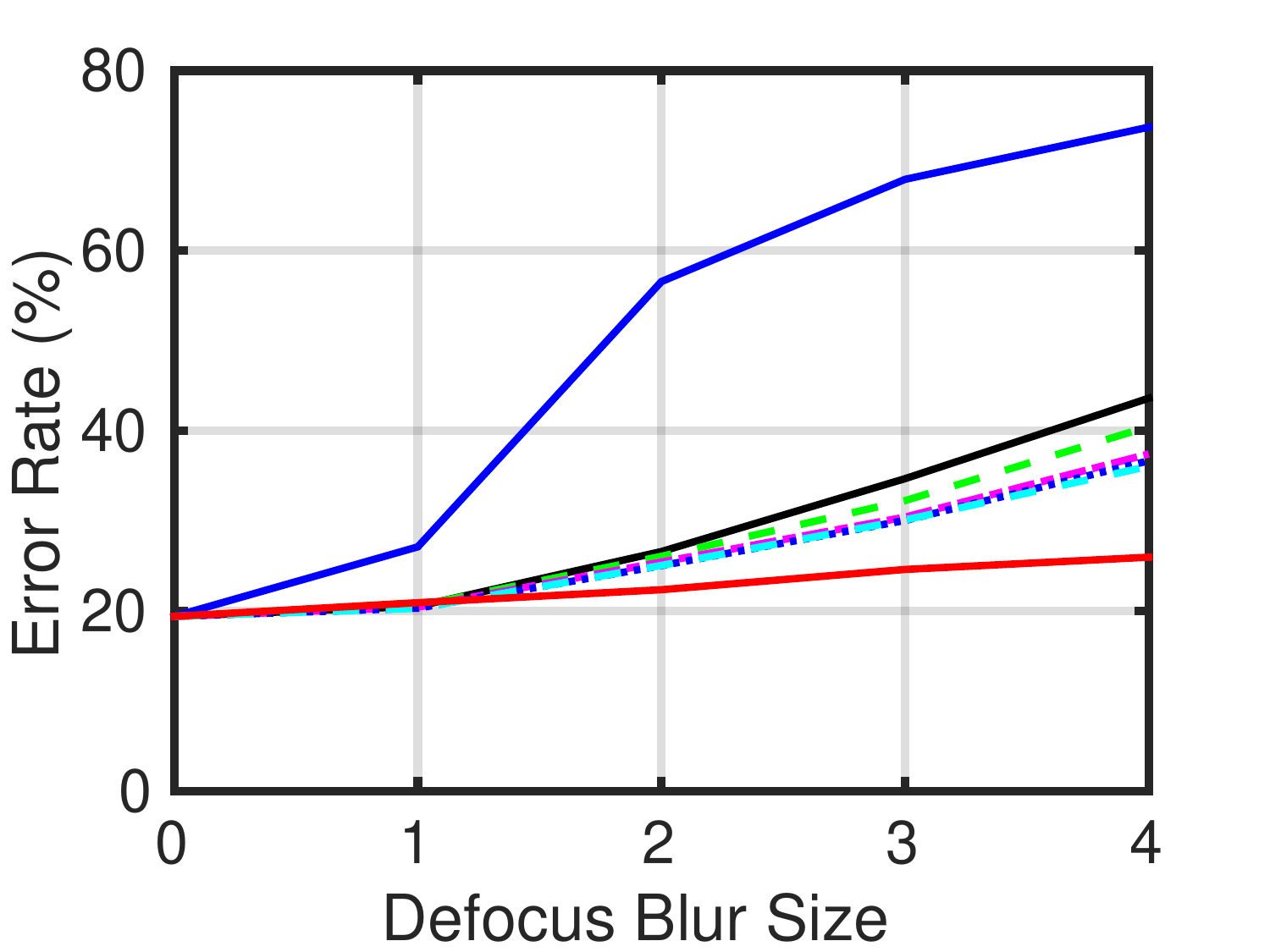}
       \label{fig:cifar-defocus}
}
\subfigure[]{
      \includegraphics[width=0.45\columnwidth]{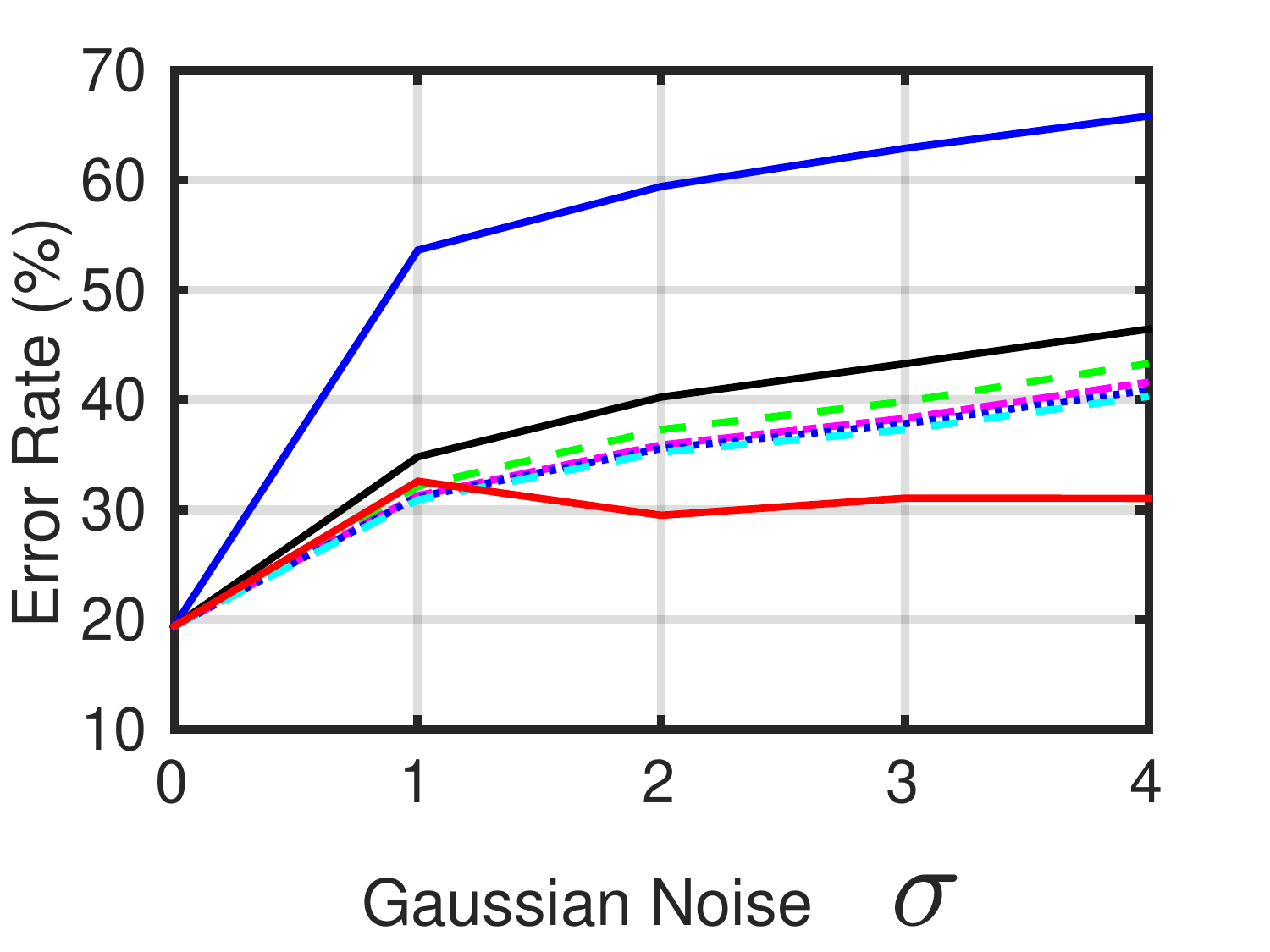}
       \label{fig:cifar-gaussian}
}
\subfigure[]{
      \includegraphics[width=0.45\columnwidth]{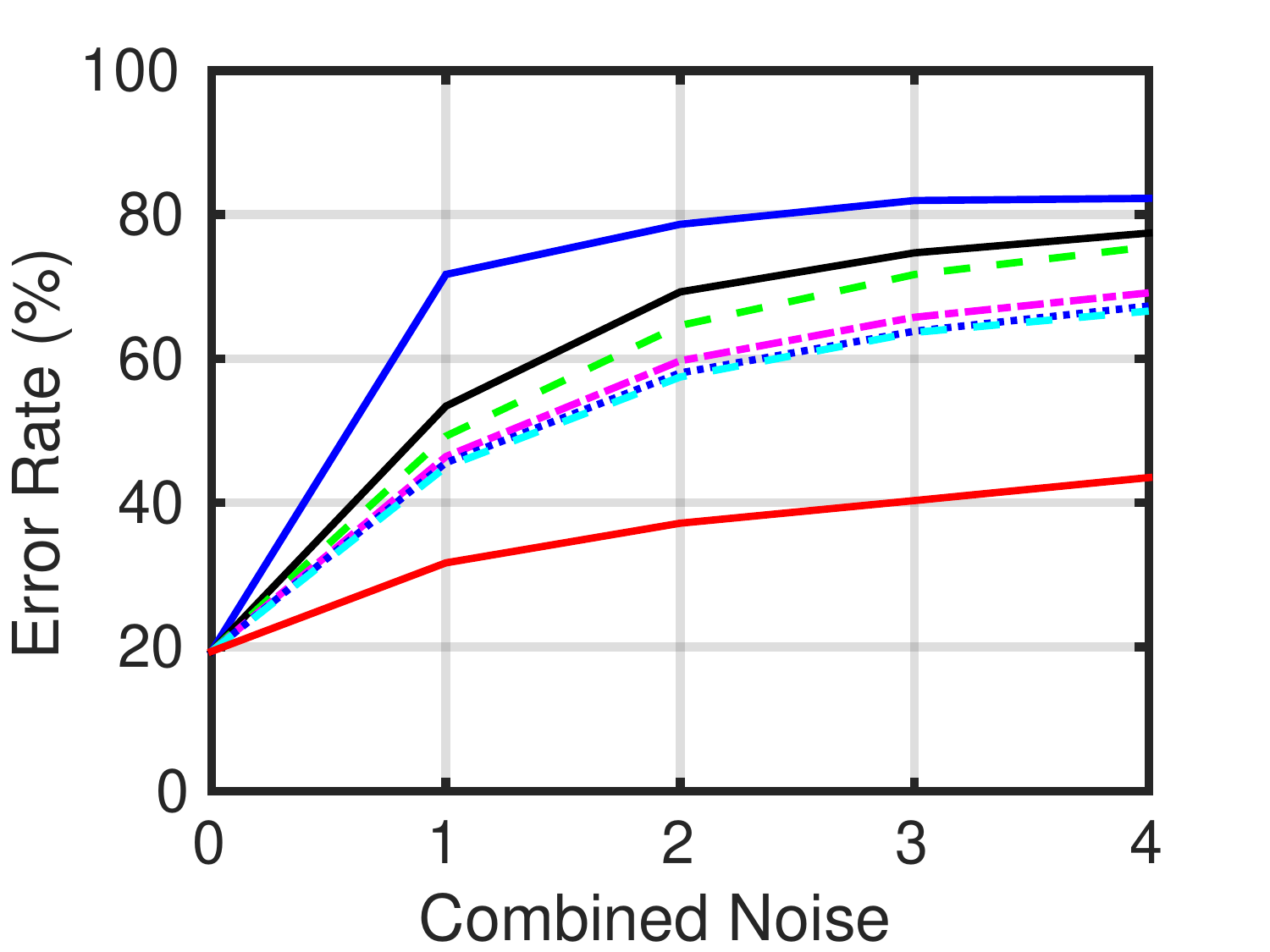}
       \label{fig:cifar-all}
}\\
\vsFigureMiddle
\caption[]{Error rates for CIFAR10-quick model on CIFAR dataset under different blurs and noises.}
\label{fig:err-cifar}
\vsFigureDown
\end{figure*}

Fig.~\ref{fig:example-mnist} and~\ref{fig:example-cifar} show the example images of blur and noise effects in MNIST and CIFAR-10, respectively. Each row of images represents one type of distortion. For the first 3 rows, only one type of distortion is applied, and for the last row, we apply all 3 types of distortion on one single image. As we see each row from left to right, the distortion level increases from 0 to 4.


Fig.~\ref{fig:example-imagenet} shows an example in ILSVRC2012 validation set. 
When we generate the distorted dataset, each image in training and testing set has random distortion values sampled from $[0,4]$ for all 3 types of distortion.

\vsSectionUp
\section{Experimental results and Analysis}
\vsSectionDown

Fig.~\ref{fig:err-mnist} and~\ref{fig:err-cifar} show the results of our experiment. We compare 3 methods: \textit{no train} means that the model is trained on the clean dataset, while tested on the noisy dataset. \textit{first-N} means that we fine-tuning the first N layers while fixing the remaining layers in the network. For LeNet-5 network, there are 4 trainable layers, so we have \textit{first-1} to \textit{first-4}, for CIFAR10-quick network, we have \textit{first-1} to \textit{first-5}.

\textbf{Results on MNIST:}
Fig.~\ref{fig:err-mnist} shows the results on MNIST dataset. 
For motion blur and Gaussian noise, the effect of distortion is relatively small (note that the scales of different plots are different). Defocus blur and combined noise have more effect on error rate. This result is consistent with the observation on Fig.~\ref{fig:example-mnist}, that the motion blur and Gaussian noise images are more recognizable than defocus blur and combined noise. MNIST dataset contains greyscale images with handwritten strokes, so edges along the strokes are important features. In our experiment, the stroke after defocus blur covers a wider area, while weakens the edge information. The motion blur also weakens edge information, but not as severe as defocus blur. This is because, under the same parameter, the area of motion blur is smaller than the defocus blur. Gaussian noise has limited effect on the edge information, so the error rate has little increase. Combined noise have much impact on the error rate.

Both fine-tuning and re-training methods can significantly reduce error rate. \textit{first-3} and \textit{first-4} have very similar results, indicating that distortion has little effect on the last several layers. When the distortion is small, fine-tuning by \textit{first-3} and \textit{first-4} achieve comparable results with re-training. When the distortion level increases, re-training achieves a better result.

\textbf{Results on CIFAR-10:}
From Fig.~\ref{fig:example-cifar} we can see the distortions in CIFAR-10 not only affect the edge information, but also have effect on color and texture information. Therefore, all 3 types of distortion can make the images difficult to recognize. This is consistent with the results shown in Fig.~\ref{fig:err-cifar}. Different from the results on MNIST dataset, all 3 types of distortion significantly worsen the  error rate on \textit{no train} result.

Using both fine-tuning and re-training methods can significantly reduce the error rate. \textit{first-3} to \textit{first-5} give similar results, indicating that the distortion mainly affects the first 3 layers. When the distortion level is low, fine-tuning and re-training have similar results. However, when the distortion level is high or under combined noise, re-training has better results than fine-tuning.

From both figures we can observe that when we fine-tune the first 3 layers, the results are very similar to fine-tuning the whole networks. This result indicates that image distortion has more effect on the low-level information of the image, while it has little effect on high-level information.

\textbf{Analysis:}
To gain some insight into the effectiveness of fine-tuning and re-training on distorted data, we look into the statistics of the feature map inside the model. Inspired by~\cite{zhang2016gmvp}, we find the mean variance of image gradient magnitude to be a useful feature. Instead of calculating the image gradient, we calculate the feature map gradient. Then, we calculate the mean variance of feature map gradient magnitude.

Given a feature map $fm$ as input, we first calculate gradient along horizontal (x) and vertical directions using Sobel filters

\vsEquationUp
\begin{equation}
s_x=\frac{1}{4}\bigl(\begin{smallmatrix}
1 & 0 & -1 \\ 
2 & 0 & -2 \\ 
1 & 0 & -1
\end{smallmatrix}\bigr), 
s_y=\frac{1}{4}\bigl(\begin{smallmatrix}
1 & 2 & 1 \\ 
0 & 0 & 0 \\ 
-1 & -2 & -1
\end{smallmatrix}\bigr)
\end{equation}
\vsEquationDown

Then we have gradient magnitude of $fm$ at location $(m, n)$ as

\vsEquationUp
\begin{equation}
g_{fm}(m, n)=\sqrt{(fm\otimes s_x)^2(m, n)+(fm\otimes s_y)^2(m, n)}
\end{equation}
\vsEquationDown

After we have the gradient magnitude $g_{fm}$ for feature map $fm$, we calculate the variance of gradient magnitude: $v_{fm}=var(g_{fm})$.

When we apply defocus blur or motion blur on an image, the clear edges are smeared out into smooth edges, thus the gradient magnitude map becomes smooth, and has lower variance. Feature maps with higher gradient variance value $v_{fm}$ are considered to have more edge and texture information, thus more helpful for image representation. While lower $v_{fm}$ value indicates that the information inside the feature map is limited, thus not sufficient for image representation.

\begin{figure}[!t]
\vsFigureUp
\subfigure[]{
      \includegraphics[width=0.48\columnwidth]{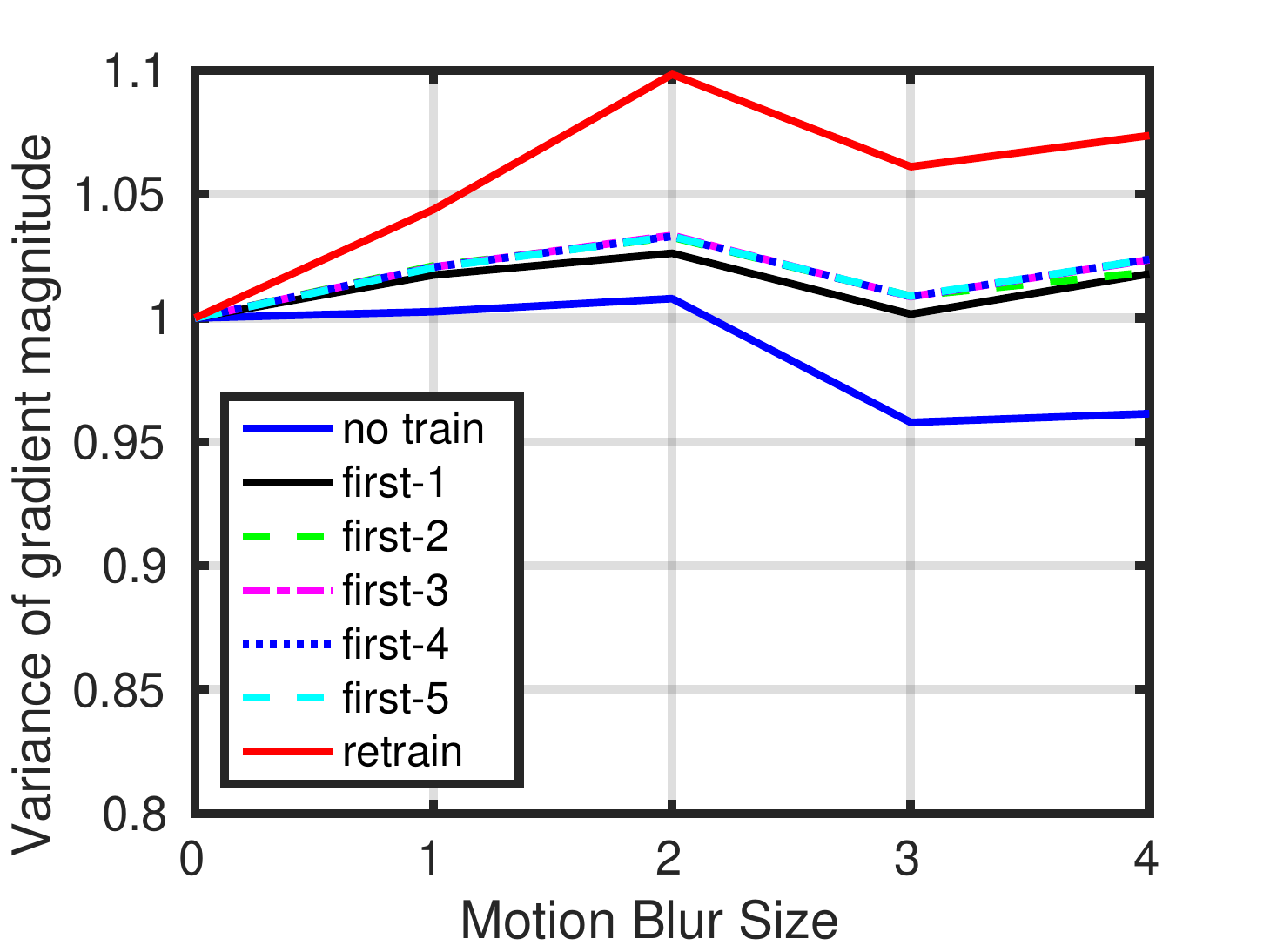}
       \label{fig:magvar-motion}
}
\subfigure[]{
      \includegraphics[width=0.48\columnwidth]{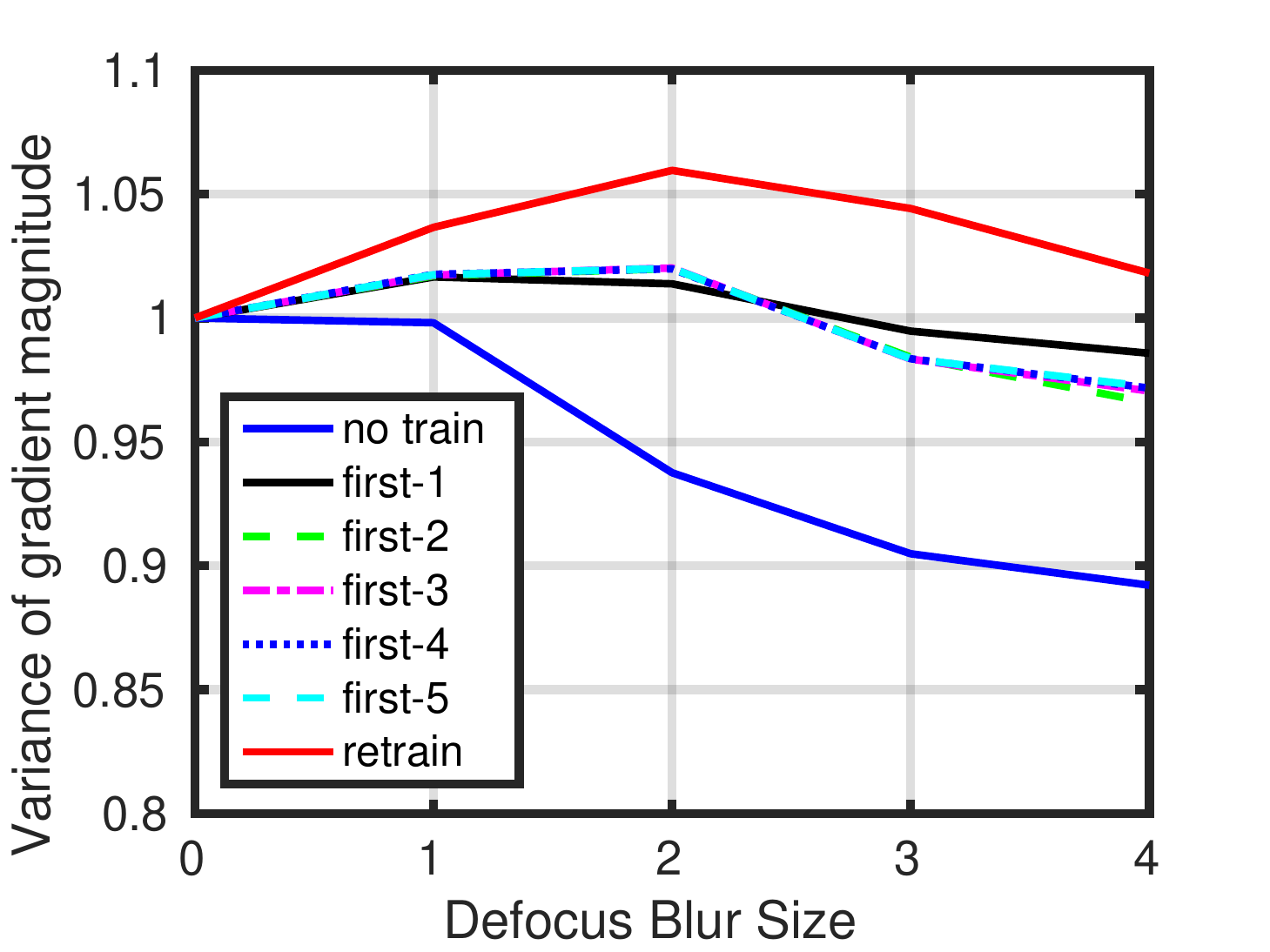}
       \label{fig:magvar-defocus}
}
\\
\vsFigureMiddle
\caption[]{Mean variance of feature map gradient magnitude for conv layer 3 of CIFAR10-quick model. (a): motion-blur. (b): defocus blur.}
\label{fig:magvar}
\vsFigureDown
\end{figure}

Fig.~\ref{fig:magvar} shows the mean variance of feature map gradient magnitude for conv3 layer(the last conv layer) of the CIFAR10-quick model. From the two figures we observe that: (1) When applying original model on distorted images, the mean variance decreases compared to applying the model on original images (see \textit{no train}), suggesting that edge or texture information is lost because of the distortion. (2) When applying fine-tuning method to the distorted images, the mean variance maintains similar as that of original images, suggesting that by fine-tuning on distorted images, the model can extract useful information from the distorted images. (3) When applying retraining method on distorted images, the mean variance is higher than applying the model on original images. It means that the retrained model fits the distorted image dataset. These results suggest that when we fine-tune the model on distorted images, we try to make the feature map representation of distorted images close to original images, so that the classification results on distorted images can be close to the results on original images. When we retrain the model on distorted images, we try to fit the DNN model on distorted dataset, and the feature map representation is not necessarily close to the representation of original images.



\begin{table}[!htbp]
\vsTableUp
\centering
\vsTableMiddle
\caption{Accuracy comparison between pre-trained Alexnet model and fine-tuned model on ImageNet validation set.}
\begin{tabular}{|c|c|c|c|c|}
\cline{1-5}
\multirow{2}{*}{} & \multicolumn{2}{c|}{original model} & \multicolumn{2}{c|}{fine-tuned model} \\
\cline{2-5}
error rate~(\%)    & clean        & distorted     & clean     & distorted     \\
                & data         &  data         &  data     &  data         \\
\cline{1-5}
top-1 error     & 42.9        & 53.1             & 42.9         & 47.7             \\
\cline{1-5}
top-5 error     & 20.1         & 28.2             & 20.4         & 23.6             \\
\cline{1-5}            
\end{tabular}
\label{tab:acc-imagenet}
\vsTableDown
\end{table}


\textbf{Results on Imagenet:}
We also examine the efficiency of fine-tuning on a large dataset and a very deep network.
For experiment on the training and validation set of ILSVRC2012, we generated the distorted data by combining all 3 types of blur/noise. For each image, and for each type of distortion, the distortion level is uniformly sampled from $[0,4]$. After obtaining the distorted data, we fine-tune the first 3 layers of a pre-trained Alexnet model~\cite{krizhevsky2012imagenet}.
Table~\ref{tab:acc-imagenet} shows the accuracy comparison between the original pre-trained Alexnet model and the fine-tuned model. Compared with the original pre-trained model, the fine-tuned model increases the performance on distorted data, while keeping the performance on clean data. When we want to use a large DNN model like Alexnet on a limited and distorted dataset, fine-tuning on first few layers can increase model accuracy on distorted data, while maintaining the accuracy of clean data.
%
\vsSectionUp
\section{Conclusions}
\vsSectionDown

Fine-tuning and re-training the model using noisy data can increase the model performance on distorted data, and re-training method usually achieves comparable or better accuracy than fine-tuning. However, there are issues we need to consider:
\begin{itemize}
    \item The size of the distorted dataset: If the model is very deep and the size of distorted dataset is small, training the model on the limited dataset would lead to overfitting. In this case, we can fine-tune the model by first N layers while fixing the remaining layers to prevent overfitting.
    \item The distortion level of noise: When the distortion level is high, re-training on distorted data has better results. When the distortion level is low, both re-training and fine-tuning can achieve good results. And in this case, fine-tuning is preferable because it converges faster, which means less computation time, and is applicable to limited size distorted datasets.
\end{itemize}

\bibliographystyle{IEEEbib}
\bibliography{refs}

\end{document}